\documentclass[10pt, journal, twoside]{IEEEtran}
\IEEEoverridecommandlockouts
\RequirePackage{etex}

\usepackage{amssymb}
\usepackage{mathtools}
\usepackage{amsthm}
\usepackage{cases}

\usepackage[ruled,linesnumbered]{algorithm2e} 
\usepackage{algpseudocode}

\usepackage{cite}
\usepackage{hyperref} 

\usepackage[pdftex]{graphicx}
\usepackage{subfigure}
\usepackage{tikz} 
\usepackage{tikz-network}
\usetikzlibrary{positioning, shapes.geometric, fit, calc, arrows.meta}
\usetikzlibrary{chains, scopes, backgrounds, shapes, shadows, quotes}

\usepackage{newtxmath}

\usepackage{booktabs}
\usepackage{float}
\usepackage{bm}
\usepackage{comment}
\usepackage{color}
\usepackage{balance}
\usepackage{textcomp}
\usepackage{stfloats}
\usepackage{url}
\usepackage{pifont}
\usepackage{verbatim}
\usepackage[capitalize]{cleveref}
\usepackage[inline]{enumitem}
\usepackage{colortbl}

\newtheorem{lemma}{Lemma}

\begin{document}

\title{Stochastic Trajectory Optimization for Robotic Skill Acquisition From a Suboptimal Demonstration}

\author{Chenlin Ming$^{1}$, Zitong Wang$^{1}$, Boxuan Zhang$^{2}$, Zhanxiang Cao$^{3}$, Xiaoming Duan$^{1}$and Jianping He$^{1}$
\thanks{$^{1}$The Department of Automation, Shanghai Jiao Tong University, and Key Laboratory of System Control and Information Processing, Ministry of Education of China, Shanghai, China.
Email: {\tt\small \{mcl2019011457, wangzitong, xduan, jphe\}@sjtu.edu.cn. }}
\thanks{$^{2}$School of Computation, Information and Technology in Technical University of Munich (TUM).
Email: {\tt\small boxuan.zhang@tum.de}.}
\thanks{$^{3}$The Department of Computer Science and Engineering, Shanghai Jiao Tong University, Shanghai, China. Email: {\tt\small caozx1110@sjtu.edu.cn}.}
\thanks{This work was supported in part by the Natural Science Foundation of Shanghai under Grant 23ZR1428900 and the National Natural Science Foundation of China under Grant 62373247 and 62303314.}
}



\maketitle
\begin{abstract}
Learning from Demonstration (LfD) has emerged as a crucial method for robots to acquire new skills. However, when given suboptimal task trajectory demonstrations with shape characteristics reflecting human preferences but subpar dynamic attributes such as slow motion, robots not only need to mimic the behaviors but also optimize the dynamic performance. 
In this work, we leverage optimization-based methods to search for a superior-performing trajectory whose shape is similar to that of the demonstrated trajectory.
Specifically, we use Dynamic Time Warping (DTW) to quantify the difference between two trajectories and combine it with additional performance metrics, such as collision cost, to construct the cost function. Moreover, we develop a multi-policy version of the Stochastic Trajectory Optimization for Motion Planning (STOMP), called MSTOMP, which is more stable and robust to parameter changes.
To deal with the jitter in the demonstrated trajectory, we further utilize the gain-controlling method in the frequency domain to denoise the demonstration and propose a computationally more efficient metric, called Mean Square Error in the Spectrum (MSES), that measures the trajectories' differences in the frequency domain. 
We also theoretically highlight the connections between the time domain and the frequency domain methods. Finally, we verify our method in both simulation experiments and real-world experiments, showcasing its improved optimization performance and stability compared to existing methods. 
The source code can be found at \href{https://ming-bot.github.io/MSTOMP.github.io}{https://ming-bot.github.io/MSTOMP.github.io}.
\end{abstract}


\section{Introduction}\label{s-introduction}
\IEEEPARstart{I}{n} the last decades, optimization-based methods have been widely used to enhance trajectory performance in high-dimensional configuration spaces. They gain great reputations for their interpretability and ease of implementation~\cite{khatib1986real,warren1989global,brock2002elastic}. 
Learning from demonstration (LfD) has also been extensively studied in recent years~\cite{ZS:2020}. Human teachers can easily transfer their skills to robots by providing high-quality demonstrations~\cite{wang2020framework}. 
However, in practice, demonstrations often turn out to be imperfect or suboptimal, and at times, even bad~\cite{ILfromImperfectDemo},~\cite{chen2021learning}. For example, gathering demonstrations involving swift motions demands a sophisticated motion capture system which is costly and requires substantial expertise. 
Looking back to the human learning process, individuals often learn new skills by imitating the experts and gradually developing their proficiency. They might follow the slow motion of the expert demonstration at the beginning and try to improve their movements through repeated practice. 
Drawing inspiration from it, we divide this learning style into two main tasks: demonstration shape imitation and motion trajectory optimization. The slow motion trajectory contains the same amount of shape characteristic information as the optimal trajectory and is better suited for kinesthetic teaching or teleoperation~\cite{ZS:2020}. By imitating the input demonstration, robots can align with human preferences~\cite{10611475} and finish the task if given enough time. 
Furthermore, motion trajectory optimization aims to improve the dynamic performance while maintaining the shape properties. 
This paper proposes an optimization-based framework that achieves imitation and optimization simultaneously.


In this paper, we propose a multi-policy version of STOMP\cite{Stomp}, called MSTOMP, to enhance the stability and exploration capability of STOMP. We extend the iterative trajectory into multiple trajectories and reconstruct the iterative process. Compared with STOMP, MSTOMP has better performance in terms of the amount of used random noise and is robust to parameter changes.
Since the stochastic trajectory optimization method does not require explicit gradient information, MSTOMP can effectively deal with complicated optimization objectives, including cases where gradients cannot be easily computed. 
To preserve the overall shape characteristic of the demonstration during the stochastic optimization process, we combine the DTW metric with other trajectory metrics, e.g., collision cost. We also propose a new metric MSES in the frequency domain which can be computed faster and has a close relationship with the DTW in the time domain. 
Finally, we demonstrate the efficiency of our method through extensive experiments in both the PyBullet simulator and the real world. 
Our contributions mainly lie in three aspects:
\begin{itemize}
    \item{We present a multi-policy Stochastic Trajectory Optimization for Motion Planning (MSTOMP), a method that extends iterative trajectories and restructures the iterative process of STOMP. This enhancement notably boosts both stability and optimization efficiency.}
    \item{We employ two metrics—Dynamic Time Warping (DTW) and Mean Squared Error (MSE) in the spectrum to measure the similarity of trajectories in time and frequency domains, respectively. Furthermore, we derive theoretical relationships between these metrics.}
    \item{We compare our method with several baselines in the PyBullet simulator~\cite{coumans2021} to demonstrate the superiorities of our framework. We also provide specific denoising methods to enhance the demonstration's quality. We implement our framework on a real-world robotic arm, verifying its outstanding capability to perform imitation and optimization simultaneously.}
\end{itemize}

Notation: We use slim symbols to denote scalars and bold symbols to denote vectors and matrices.
Let $\bm{I}_N$ be an identity matrix with dimension $N\times N$.
The sets of real numbers and complex numbers are denoted by $\mathbb{R}$ and $\mathbb{C}$, respectively. 
Let $\lvert \cdot \rvert$ denote the modulus of a vector.

The remainder of this paper is organized as follows: Section~\ref{s-relatedwork} provides essential related work of trajectory optimization and imitation from demonstration.
Section~\ref{s-methodology} presents the problem formulation and the details of the proposed MSTOMP framework.
Section~\ref{s-simulation} shows the experiments and the results of the algorithm. Finally, Section~\ref{s-conclusion} concludes the work.

\section{Related Work}\label{s-relatedwork}
LfD methods have been widely used to transfer human knowledge to robots\cite{si2021review,li2022learning}. Many works try to learn movement primitives from human demonstrations. Dynamic movement primitives (DMP)\cite{DMP} describe the trajectory as an attractor dynamics model with a learnable autonomous forcing term where the basis functions must be chosen carefully. Besides, probability approaches such as Gaussian mixture models\cite{GMM}, kernelized movement primitives\cite{KMP}, and probabilistic movement primitives\cite{ProMP} model demonstrations as probability distributions and allow robots to learn from multiple trajectories. Although these models can learn more demonstration information, requiring multiple demonstrations also makes the data acquisition and processing more complicated compared to DMP. The same disadvantage exists in the data-driven approaches. For example, GAIL\cite{GAIL} needs sufficient observation data and is hard to use beyond the training environment. Moreover, an alignment procedure must be in place to process multiple trajectories, and the performance is easily influenced by imperfect trajectories\cite{xu2024learning}.

Optimization-based methods primarily tackle the motion planning task by formulating an optimization problem. CHOMP\cite{CHOMP} presents a functional gradient approach to optimize motion planning by iteratively reducing the value of a cost function, which integrates both trajectory smoothness and collision avoidance into a single framework. TrajOpt\cite{TrajOpt} presents sequential convex optimization to solve motion planning problems, explicitly incorporating collision avoidance constraints into the optimization process. STOMP~\cite{Stomp} focuses on generating optimal robot trajectories through stochastic optimization, using noises to search for feasible trajectories effectively. Recently, Inverse reinforcement learning~\cite{chen2021learning} and other information-theoretic methods, like Model Predictive Path Integral (MPPI)~\cite{InformationMPC}, have shown great potential for finding near-optimal reward functions or policies. Optimization-based methods can polish suboptimal trajectories by optimizing a pre-defined cost function. However, they often struggle to capture the human preferences presented in demonstrations which are important in motion planning tasks.

Koert~\cite{Demonstrationtraj} and Osa~\cite{guidingosa2017} propose methods combining optimization-based approaches and LfD approaches. By learning the distribution of trajectories demonstrated by human experts, they generate trajectories that maintain demonstrated behaviors while adapting to new situations, such as avoiding newly introduced obstacles. Later, Osa et al.\cite{osa2020multimodal} introduce a stochastic multimodal trajectory optimization algorithm that determines multiple trajectory solutions by identifying the modes of the cost function and optimizing trajectories corresponding to these modes. However, acquiring movement demonstrations is often expensive. In this paper, we focus on learning from a single suboptimal demonstration. We relax the requirement on the perfect dynamic performance of the demonstration, merely requiring that the demonstration shape encapsulates human prior knowledge, e.g., slow motion. By combining optimization-based methods and LfD, we empower robots to acquire new skills from a suboptimal demonstration.

\section{Methodology}\label{s-methodology}
This section mainly introduces the overall theoretical framework and the details of the implementation of the MSTOMP algorithm. To equip the MSTOMP algorithm with the capacity to imitate demonstrations, we introduce two metrics, i.e., Dynamic Time Warping (DTW) and Mean Square Error in the Spectrum (MSES), to evaluate the similarity between trajectories in the time and frequency domains, respectively.

\subsection{Problem Formulation}\label{ss-3.1}
We define all trajectories as discrete point sequences. A trajectory can be represented by $\bm{\theta} \in \mathbb{R}^{M \times N}$, where $N$ denotes the length of the discrete sequence and $M$ represents the dimension of the points. For example, for a 7-DOF robotic arm, the joint position trajectory in its configuration space can be represented by $\bm{\theta} \in \mathbb{R}^{7 \times N}$.
We assume that a robot has a fixed maximum control frequency $f$, which determines the time interval between two points on a discrete trajectory~$\bm{\theta}$. 
Under this assumption, we know the motion duration $T = (N-1)/f$ from the length of the discrete sequence. 
We then define $\tilde{\bm{\theta}}$ as a noisy joint position trajectory following a normal distribution $\tilde{\bm{\theta}}_i \sim \mathcal{N}(\bm{\theta}_i, \bm{\Sigma}), i = 1,2,\dots,N$ with mean $\bm{\theta}_i$ and covariance~$\bm{\Sigma}$. The cost function of the noisy trajectory $\tilde{\bm{\theta}}$ is defined as:
\begin{equation}\label{eq: QQQ}
    Q(\tilde{\bm{\theta}}) = \sum_{i=1}^N q(\tilde{\bm{\theta}}_i) + \frac{1}{2} \sum_{i=1}^N{\tilde{\bm{\theta}_i}}^\top \bm{R} \tilde{\bm{\theta}_i},
\end{equation}
where $\bm{R}$ is a positive semi-definite matrix. The $\sum_{i=1}^N q(\tilde{\bm{\theta}}_i)$ term is a combination of arbitrary state-dependent cost terms including collision costs, expected velocity costs, and other specific costs. The $\frac{1}{2} \sum_{i=1}^N\tilde{\bm{\theta}_i}^\top \bm{R} \tilde{\bm{\theta}_i}$ term represents the smoothness control cost.

Based on \eqref{eq: QQQ}, our optimization problem is formulated as:
\begin{equation}\label{eq: formulation}
    \min_{\bm{\theta}} \mathbb{E} \left[ \sum_{i=1}^N q(\tilde{\bm{\theta}}_i) + \frac{1}{2} \sum_{i=1}^N{\tilde{\bm{\theta}_i}}^\top \bm{R} \tilde{\bm{\theta}_i}\right].
\end{equation}
We compute the gradient of the expectation in \eqref{eq: formulation} with respect to $\bm{\theta}$ and obtain:
\begin{equation}
\begin{aligned}
    \nabla_{\bm{\theta}} \left( \mathbb{E}\left[ \sum_{i=1}^N q ( \tilde{\bm{\theta}}_i) + \frac{1}{2} \sum_{i=1}^N{\tilde{\bm{\theta}_i}}^\top \bm{R} \tilde{\bm{\theta}_i} \right] \right) = 0\\
    \Rightarrow \quad \mathbb{E} [ \tilde{\bm{\theta}} ] = -\bm{R}^{-1} \mathbb{E} \left[ \nabla_{\bm{\theta}}
    \left( \sum_{i=1}^N q ( \tilde{\bm{\theta}}_i) \right) \right].
\end{aligned}
\end{equation}
Note that the complex composition of the cost function may render challenges in calculating gradients. As in \cite{Stomp,theodorou2010reinforcement}, we estimate the gradient using the generated noisy trajectories: 
\begin{equation}\label{eq: expectation of theta}
\begin{aligned}
    \mathbb{E} [ \tilde{\bm{\theta}} ] & = -\bm{R}^{-1} \left[\int \delta \tilde{\bm{\theta}} d \bm{P} \right]\\
    & = -\bm{R}^{-1} \left[\int{\exp \left(-\frac{1}{\lambda} \bm{S} (\tilde{\bm{\theta}})\right) \delta \tilde{\bm{\theta}} d (\delta\tilde{\bm{\theta}})}\right],
\end{aligned}
\end{equation}
where the probability measure $\bm{P}$ represents $\exp \left(-\frac{1}{\lambda} \bm{S} (\tilde{\bm{\theta}}) \right)$ and the state-dependent cost is defined as $\bm{S}(\tilde{\bm{\theta}}) = \sum_{i=1}^{N} q ( \tilde{\bm{\theta}}_i)$. Intuitively, a noisy trajectory with a lower cost should have a higher probability of influencing the direction of optimization. Using the equations above, $\mathbb{E}\left[Q(\tilde{\bm{\theta}})\right]$ can be minimized by leveraging generated noisy trajectories. 

\subsection{Implementation Details of the MSTOMP} \label{ss-3.2}
\setlength{\textfloatsep}{10pt}
\begin{algorithm}[t]
\caption{MSTOMP Algorithm} \label{alg: MSTOMP}
\KwIn{An initial trajectory vector $\bm{\theta}_{init} \in \mathbb{R}^{7 \times N}$;}
\KwOut{Optimized trajectory $\bm{\theta}_b \in \mathbb{R}^{7 \times N}$;}
Predefine: A state-dependent cost function $q:\mathbb{R}^7 \rightarrow \mathbb{R}$, a given matrix $\bm{R} \in \mathbb{R}^{N \times N}$;\\
Initialize iterative trajectories $\bm{\theta}_{b}, \bm{\theta}_{d}, \bm{\theta}_{p} = \bm{\theta}_{init}$;\\
Initialize reused trajectory set: $\{\bm{\theta}_{reused,i}\}_{i=1}^n$;\\
\While{Convergence of $Q(\bm{\theta}_b)$ in \eqref{eq: QQQ} is not achieved}{
    Create $K$ noisy sequences $\bm{\epsilon}_1, \ldots, \bm{\epsilon}_K$, where $\bm{\epsilon}_k \in \mathbb{R}^{7 \times N}, \left[\bm{\epsilon}_k^\top\right]_j \sim \mathcal{N}(\bm{0}, \bm{R}^{-1}), k = 1, \ldots, K, j = 1, \ldots, 7$;\label{algline:addnoise-start}\\
    \For{$\bm{\theta}$ in $\left\{\bm{\theta}_{d}, \bm{\theta}_{p}\right\}$}{
        Create $K$ noisy trajectories $\bm{\tilde{\theta}}^1, \ldots, \bm{\tilde{\theta}}^K$ with each $\bm{\tilde{\theta}}^k = \bm{\theta} + \bm{\epsilon}_k$;\\
    \For{$k=1, \ldots, K$}{
        Compute $\bm{S}(\bm{\tilde{\theta}})_{k,j}= q([\bm{\tilde{\theta}}^{k}]_{j}), \bm{S} \in \mathbb{R}^{K \times N}$;\label{algline: S}\\
        Compute $\bm{P}(\bm{\tilde{\theta}})_{k,j} = \frac{e^{-\frac{1}{\lambda} S(\bm{\tilde{\theta}})_{k,j}}}{\sum_{k=1}^{K}[e^{-\frac{1}{\lambda} S(\bm{\tilde{\theta}})_{k,j}}]}, \bm{P} \in \mathbb{R}^{K \times N}$;\label{algline: P}\\
    }
    Replace high-cost noisy trajectories with reused trajectories;\label{algline:reusable}\\
    \For{$j=2, \ldots, N-1$}{
        Compute $\delta \bm{\theta}^\top_j = \sum_{k=1}^K \bm{P}(\bm{\tilde{\theta}})_{k,j}[\bm{\epsilon}_k^\top]_j$;\\
    }
    Update $\bm{\theta} \leftarrow \bm{\theta} + \gamma \bm{R}^{-1}\delta \bm{\theta} \in \mathbb{R}^{N \times 7}$;\label{algline:withgamma}\label{algline:addnoise-end}\\
    Compute $Q(\bm{\theta})$;\label{algline:reused-start}\\
    Compute $m = \mathop{\arg\max}\limits_{i}{Q(\bm{\theta}_{reused,i})}$;\\
    \uIf{$Q(\bm{\theta}) < Q(\bm{\theta}_{reused,m})$}{
        $\bm{\theta}_{reused,m} \leftarrow \bm{\theta}$;
    }\label{algline:reused-end}
    }
    Update $Q(\bm{\theta}_b) \leftarrow \min\{Q(\bm{\theta}_d), Q(\bm{\theta}_p)\}$;\\ 
    Update $\bm{\theta}_b \leftarrow \mathop{\arg\min}\limits_{\bm{\theta}}\{Q(\bm{\theta}_d), Q(\bm{\theta}_p)\}$;\\
    Update $\bm{\theta}_p \leftarrow \bm{\theta}_b$ at a scheduled frequency.\label{algline:update-s}
}
\end{algorithm}

In this section, we exemplify the implementation of the MSTOMP algorithm on a 7-DOF robotic arm. The trajectory is represented by a sequence of positions for the robotic joints $\bm{\theta} \in \mathbb{R}^{7 \times N}$. We design our framework based on STOMP, where $K$ noise sequences $\bm{\epsilon}_1, \ldots, \bm{\epsilon}_K$ are sampled from $\mathcal{N}(\bm{0}, \bm{R}^{-1})$ and added to the trajectory at the current iteration respectively to search for trajectories with lower costs, as shown in line \ref{algline:addnoise-start} to \ref{algline:addnoise-end} of Algorithm \ref{alg: MSTOMP}. 
However, the performance of STOMP entirely depends on the quality of the generated noise. In other words, if generated noisy trajectories all have high costs, then the trajectories gradually become worse with iterations.

Unlike previous works focusing on improving the distribution of the generated noise~\cite{mukadam2016gaussian}, we modify the iterative process by extending STOMP into a multi-policy version to overcome the above shortcomings. Specifically, we maintain three types of trajectories: the best trajectory $\bm{\theta}_b$, the distal exploration trajectory $\bm{\theta}_d$, and the proximal searching trajectory $\bm{\theta}_p$. 
The best trajectory $\bm{\theta}_b$ records the trajectory with the minimum cost encountered so far.
The distal exploration trajectory $\bm{\theta}_d$ is updated in every iteration regardless of whether the cost increases. This strategy enables the algorithm to search for the global optimal solution and avoid the local optimal solution. 
The proximal searching trajectory $\bm{\theta}_p$ is responsible for exploring possible better trajectories near the current best trajectory, 
and we periodically update $\bm{\theta}_p$ as shown in line \ref{algline:update-s} of Algorithm \ref{alg: MSTOMP} to prevent potential degradation of $\bm{\theta}_p$ due to accumulated noise.

To enhance the stability of the MSTOMP, we save low-cost trajectories and reuse them to improve the quality of generated noisy trajectories. As a detailed explanation to line \ref{algline:reusable} of Algorithm~\ref{alg: MSTOMP}, we replace high-cost noisy trajectories with reused trajectories and modify corresponding values in matrices $\bm{S}$ and $\bm{P}$, consistent with \eqref{eq: expectation of theta} and line \ref{algline: S} to \ref{algline: P}. Note that the number of reused trajectories $n$ should be strictly less than the number of noisy sequences $K$. Generating new noisy trajectories is essential to maintain the potential for encountering superior trajectories. In line~\ref{algline:withgamma}, we add a dynamic decay factor $\gamma$ to suppress the noise for better performance. In line~\ref{algline:reused-start} to \ref{algline:reused-end}, we calculate the iterative trajectory's cost and record low-cost trajectories to improve the quality of reused trajectories. We initialize the reused trajectories with infinite cost values. During the iterative process, whenever the iterative trajectory's cost becomes less than the maximum cost value in the reused trajectories, a replacement operation is executed in Line \ref{algline:reused-end}.
The details of the MSTOMP algorithm can be seen in Algorithm~\ref{alg: MSTOMP}.

\subsection{Measuring Similarity using DTW in MSTOMP}\label{ss-3.3}
The quality of imitation relies on the similarity between the optimized trajectory and the demonstration in the 3-dimensional Euclidean space.
Therefore, we use $\bm{A} = \begin{bmatrix} \bm{p}_1, \cdots, \bm{p}_N \end{bmatrix}^\top \in ~\mathbb{R}^{N \times 3}, \hat{\bm{A}} = \begin{bmatrix}\hat{\bm{p}}_1, \cdots, \hat{\bm{p}}_{\hat{N}}\end{bmatrix}^\top \in \mathbb{R}^{\hat{N} \times 3}$ to denote two motion trajectories of the end-effector in Euclidean space, where $\bm{p}_i = (x_i, y_i, z_i)^\top$ represents the $i$-th coordinate point in the end-effector's trajectory in Euclidean space. Notably, these trajectories may not be temporally aligned. We fix the start and end points as $p_1 = \hat{p}_1$ and $p_N = \hat{p}_{\hat{N}}$, respectively.
Furthermore, forward and inverse kinematics can connect joint space and task space coordinate points by $\bm{\theta} = IK(\bm{A})$ and $\bm{A} = FK(\bm{\theta})$. 

To equip the robot with the ability to imitate demonstrations, we introduce an extra term to \eqref{eq: QQQ} in the MSTOMP algorithm that computes the similarity between the current trajectory and the demonstration. We apply appropriate penalties to encourage the trajectory under optimization to imitate the demonstration. By balancing the imitation component and other optimization components, we can effectively leverage the MSTOMP to address demonstration imitation problems. In this subsection, we choose Dynamic Time Warping (DTW) as a metric to measure the similarity between discrete trajectories with different lengths in the time domain for its easy implementation and good performance. 
DTW is a dynamic programming-based algorithm widely used in language processing\cite{DTW_speech} and matching unaligned data\cite{superhumanvan2010}. It is also used for measuring trajectory similarity and aligning multiple trajectories\cite{tao2021comparative,hu2023spatio}. 
Details of the DTW can be seen in Algorithm \ref{alg: DTW}. As a special note, $d(\cdot, \cdot)$ represents a valid distance measure (e.g., Euclidean distance) that measures the distance between two vectors, which is crucial for determining the similarity between two trajectories. DTW algorithm has time complexity $\mathcal{O}(N\hat{N})$, and improved DTW algorithms such as FastDTW\cite{salvador2007toward} can be implemented in practice.
In consequence, we add an extra term $q_d(\bm{\tilde{\theta}}, \bm{\theta}_{demo}) = \text{DTW}\left(FK(\tilde{\bm{\theta}}), FK(\bm{\theta}_{demo})\right)$ into $Q(\bm{\tilde{\theta}})$ to encourage imitation of the demonstration, and the modified loss function is as follows:
\begin{align}\label{eq: extended-QQQ}
    Q(\tilde{\bm{\theta}}) & \!=\! \sum_{i=1}^N q(\tilde{\bm{\theta}}_i) \!+\! \text{DTW}\left(FK(\tilde{\bm{\theta}}), FK(\bm{\theta}_{demo})\right) \!+\! \frac{1}{2} \sum_{i=1}^N\tilde{\bm{\theta}_i}^\top \bm{R} \tilde{\bm{\theta}_i} \nonumber \\
    & = \sum_{i=1}^N \left(q(\tilde{\bm{\theta}}_i^\top) + \frac{1}{N}q_d(\tilde{\bm{\theta}}, \bm{\theta}_{demo})\right) + \frac{1}{2} \sum_{i=1}^N\tilde{\bm{\theta}_i}^\top \bm{R} \tilde{\bm{\theta}_i}.
\end{align}

\setlength{\textfloatsep}{10pt}
\begin{algorithm}[t]
\caption{DTW Algorithm} \label{alg: DTW}
\KwIn{Trajectory 1 $\bm{A} \in \mathbb{R}^{N \times 3}$, Trajectory 2 $\hat{\bm{A}} \in \mathbb{R}^{\hat{N} \times 3}$;}
\KwOut{Similarity measures $\text{DTW}(\bm{A}, \hat{\bm{A}})$;}
Initialization DTW array $\bm{S} \in \mathbb{R}^{(N+1) \times (\hat{N}+1)}$; \\
\For{$i = 0, ..., N$}{
    \For{$j = 0, ..., \hat{N}$}{
        $\bm{S}_{i,j} = \inf$;
    }
}
Assign $\bm{S}_{0, 0} = 0$;\\
\For{$i = 1, ..., N$}{
    \For{$j = 1, ..., \hat{N}$}{
        $cost = d(\bm{A}_i^\top, \bm{A}_j^\top)$;\\
        $\bm{S}_{i,j} = cost + \min(\bm{S}_{i-1,j}, \bm{S}_{i,j-1}, \bm{S}_{i-1,j-1})$;\\
    }
}
$\text{DTW}(\bm{A}, \hat{\bm{A}}) = \bm{S}_{N,\hat{N}}$.
\end{algorithm}

\subsection{Measuring Similarity with Spectrum Analysis}
\label{ss-3.4}
Since the acquired demonstrations may be noisy, in this section, we propose a similarity metric in the frequency domain to reduce the influence of noise.
Firstly, we use the Discrete Fourier Transform (DFT) to convert $\bm{A}, \hat{\bm{A}}$ in the time domain into a sequence in the frequency domain $\bm{\mathcal{F}}, \bm{\hat{\mathcal{F}}} \in \mathbb{C}^{N \times 3}$. Note that $\bm{\mathcal{F}}, \bm{\hat{\mathcal{F}}}$ have the same length because we apply zero-padding for preprocessing, which is a classic technique used in DFT to guarantee that the length of the input signal is a power of two. By leveraging Parseval's theorem \cite{parseval1806memoire}, we introduce the Mean Square Error in the Spectrum (MSES) as our trajectory similarity metric in the frequency domain and establish the theoretical relationship between DTW and MSES.
\begin{lemma}[Parseval's theorem \cite{parseval1806memoire}]\label{lemma1}
    Given $\bm{x} \in \mathbb{R}^{M \times N}$ and its DFT $\bm{X} \in \mathbb{C}^{M \times N}$, we have:
    \begin{equation}\label{eq: parseval}
        \sum_{i=0}^{M-1} \sum_{k=0}^{N-1} \lvert x_{i, k}\rvert^2 = \frac{1}{MN} \sum_{u=0}^{M-1} \sum_{v=0}^{N-1} \lvert X_{u,v}\rvert^2.
    \end{equation}
\end{lemma}

Parseval's theorem reveals the relationship between a signal in the time domain and its representation in the frequency domain. Both the sum of the squares of the time-domain signal and the sum of the squares of the magnitudes in the frequency domain represent the total energy of the signal. By using the DFT, spectrum analysis can better capture the overall properties of the signal, such as periodicity and shape information. We then introduce the extended version of Parseval's theorem to describe the relationship between two different signals in the time and frequency domains.

\begin{lemma}[Parseval's theorem {\cite[Eq. (78)]{baddour2019discrete}}]\label{2-DParsel}
    For the 2-dimensional DFT for different signals $\bm{x} \in \mathbb{R}^{M \times N}$ and $\bm{y} \in \mathbb{R}^{M \times N}$, we have:
    \begin{equation}\label{eq: parseval-2}
        \sum_{i=0}^{M-1} \sum_{k=0}^{N-1} x_{i,k} y_{i,k} = \frac{1}{MN} \sum_{u=0}^{M-1} \sum_{v=0}^{N-1} X_{u,v} Y^*_{u,v},
    \end{equation}
    where $\bm{X} \in \mathbb{C}^{M \times N}$ is the DFT of $\bm{x}$, $\bm{Y} \in \mathbb{C}^{M \times N}$ is the DFT of $\bm{y}$, and $*$ represents the complex conjugate.
\end{lemma}

In our problem, if we can minimize the difference between $\bm{\mathcal{F}}$ and $\bm{\hat{\mathcal{F}}}$ in the frequency domain, then the trajectories $\bm{A}, \hat{\bm{A}}$ should also show some similarity in the time domain.
We take the Mean Square Error (MSE) as a metric to calculate the difference between $\bm{\mathcal{F}}$ and $\bm{\hat{\mathcal{F}}}$. 
The following lemma shows that minimizing the MSE in the spectrum (MSES) in the frequency domain is consistent with minimizing the DTW in the time domain.

\begin{lemma}[]\label{lmm:DTW<=MSE}
    For different trajectories after zero-padding $\bm{A}, \hat{\bm{A}} \in \mathbb{R}^{N \times 3}$ and their spectrum representations $\bm{\mathcal{F}}, \bm{\hat{\mathcal{F}}} \in \mathbb{C}^{N \times 3}$, we have:
    \begin{align}
        \mathrm{DTW}(\bm{A}, \hat{\bm{A}}) & \leq \mathrm{MSES}(\bm{\mathcal{F}}, \bm{\hat{\mathcal{F}}})\label{eq: dtw-mse}\\
        & = \frac{1}{3N}\sum_{u=0}^{N-1}\sum_{v=0}^{2}(f_{u,v} - \hat{f}_{u,v})(f_{u,v} - \hat{f}_{u,v})^*.\label{eq: mses}
    \end{align}
\end{lemma}

Lemma \ref{lmm:DTW<=MSE} is established by utilizing Lemma \ref{2-DParsel} to convert the frequency-domain MSE into the time-domain MSE of zero-padded, equal-length trajectories. By the fact that DTW is less than or equal to the MSE value in the time domain, we arrive at the final inequality~\eqref{eq: dtw-mse}. Lemma \ref{lmm:DTW<=MSE} indicates that the $\mathrm{MSES}(\bm{\mathcal{F}}, \hat{\bm{\mathcal{F}}})$ is an upper bound for $\text{DTW}(\bm{A}, \hat{\bm{A}})$, and we can minimize the MSES as a surrogate for the DTW. By using MSES, we can quantify the trajectory similarity after denoising in the frequency domain without the need to transform them back to the time domain. Moreover, the time complexity of MSES is $\mathcal{O}(N)$, which means faster computation and a shorter optimization process. Note that the MSES is only one method to measure the difference between $\bm{\mathcal{F}}$ and $\hat{\bm{\mathcal{F}}}$. Other approaches for calculating the difference can also be employed. 
Parseval's theorem establishes a connection between the signals in frequency and time domains, providing the quadratic indicators in the frequency domain with enhanced interpretability.

\begin{figure}
    \centering
    \includegraphics[width=0.9\linewidth]{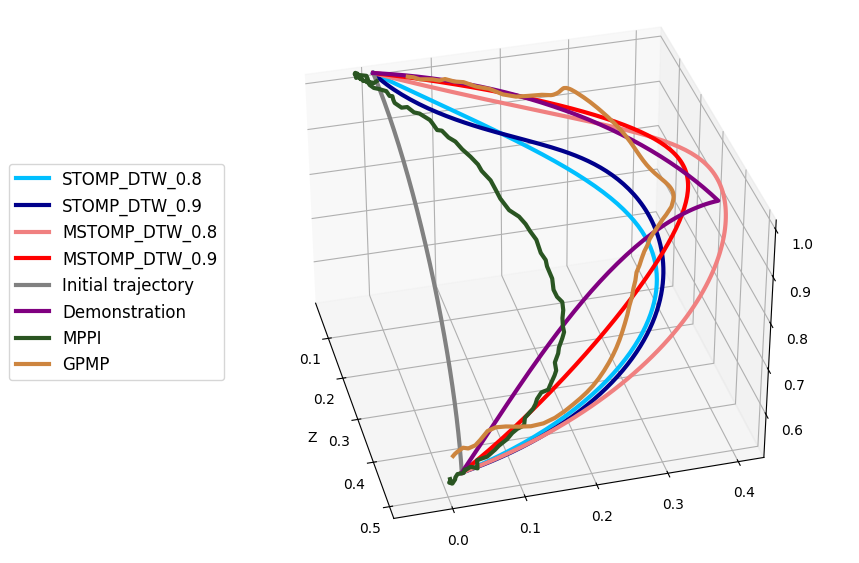}
    \vspace{-1em}
    \caption{Examples of optimized trajectories using STOMP, MSTOMP, MPPI, and GPMP. All algorithms utilize the DTW as the cost function. The number in each legend label is the value of the decay factor $\gamma$ used in the algorithms.}
    \label{stomp_mstomp_trajectories}
    \vspace{-1em}
\end{figure}

\begin{table}[htbp]
\centering
\caption{DTW Trajectory Similarities}
\begin{tabular}{l|c|l|c}
\toprule
Methods & Values & Methods & Values\\
\midrule
MPPI & 18.68 & GPMP & 6.10\\
STOMP ($\gamma$=0.8)& 13.45 & STOMP ($\gamma$=0.9)& 10.31\\
MSTOMP ($\gamma$=0.8) & 7.8 & MSTOMP ($\gamma$=0.9) & \textbf{4.07}\\
\bottomrule
\end{tabular}
\label{tab: DTW loss}
\vspace{-1em}
\arrayrulecolor{black}
\end{table}

\section{Experiments on a Robotic Arm} \label{s-simulation}
In this section, we verify the effectiveness of the MSTOMP algorithm in both simulation and real-world experiments. To show that our algorithm is competent for solving high-dimensional trajectory optimization problems, we first conduct simulation experiments on a 7-DOF Panda manipulator in the PyBullet simulator \cite{coumans2021}. We compare our algorithm with STOMP and show its stability and exploration capability. All the simulation results are produced on a computer with an RTX1650 GPU and an Intel-i7 CPU. Furthermore, we conduct real-world experiments on a 6-DOF Unitree Z1 pro robotic arm \cite{unitreeZ1} and demonstrate the performance of the MSTOMP algorithm in practice.

\subsection{Performance Comparison with Baselines}
To demonstrate the effectiveness of MSTOMP, a simulation experiment is designed where a robotic arm needs to imitate a demonstration trajectory. We implement STOMP\cite{Stomp}, MPPI\cite{InformationMPC}, and Gaussian Process Motion Planning (GPMP)~\cite{mukadam2016gaussian} as our baselines.
In all experiments, DTW is used as the imitation term, and the final optimized trajectories are shown in \cref{stomp_mstomp_trajectories}.
The demonstration trajectory is visualized in purple, the initial trajectory in gray, and other colors represent the trajectories optimized by different baselines.
To quantitatively compare the results, we calculate the DTW values between the optimized trajectories and the demonstration trajectories for all methods, and they are reported in~Table~\ref{tab: DTW loss}. The results demonstrate that MSTOMP has better capability in minimizing DTW values. Specifically, MSTOMP with $\gamma=0.9$ achieves the best imitation scores overall, significantly outperforming MPPI. Note that GPMP also yields a relatively low DTW value. However, as shown in~\cref{stomp_mstomp_trajectories}, the trajectory produced by GPMP exhibits greater irregularities than those produced by MSTOMP, despite their comparable numerical performance.

\begin{figure}
    \centering
    \includegraphics[width=0.9\linewidth]{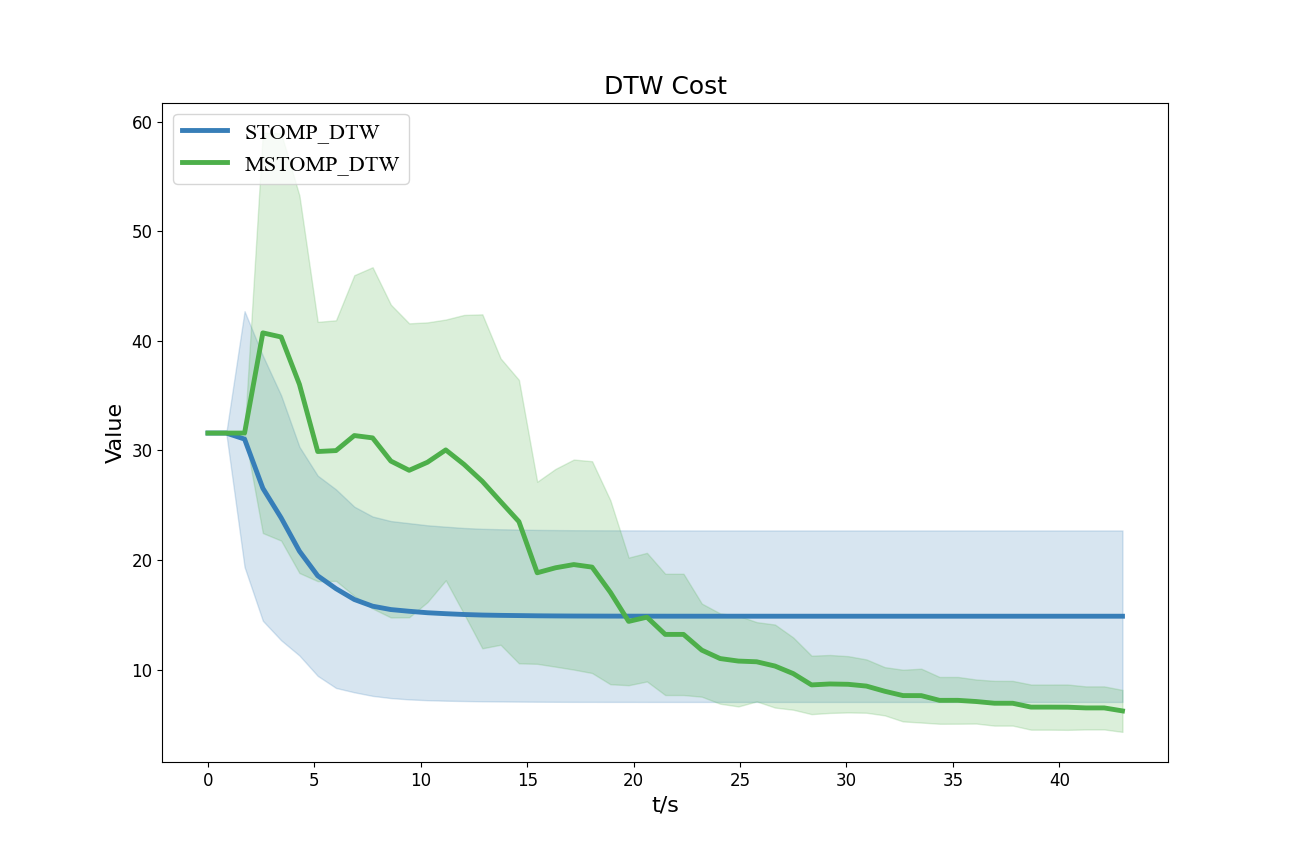}
    \vspace{-0.5cm}
    \caption{The evolution of the DTW value over $100$ repeated experiments. MSTOMP has a lower variance and finds better trajectories. At the end of the optimization process, STOMP exhibits a standard deviation of 7.6, whereas MSTOMP achieves a significantly lower standard deviation of 1.4.}
    \label{stomp_mstomp_dtw_cost}
    \vspace{-1em}
\end{figure}

Note that stochastic trajectory optimization is subject to variability in its convergence path, which is significantly affected by the quality of the intermediate noise-generated trajectories.
To reduce the influence of randomness and quantitatively evaluate the imitation performance of STOMP and MSTOMP, we conduct the experiment $100$ times for each method using the same set of parameters ($\gamma=0.9$, $K=20$ and so on). The DTW values during the optimization process are plotted in~\cref{stomp_mstomp_dtw_cost}.
Given that MSTOMP requires additional time and calculations for the same number of iterations, we extend the duration for STOMP to run for a commensurate period, nearly doubling its number of iterations. Due to the advantages of the multi-policy structure in our MSTOMP method, the corresponding DTW value ultimately converges to a smaller mean, showing its outperforming exploratory capabilities. Simultaneously, the regular updating of $\bm{\theta}_d$ and $\bm{\theta}_b$ results in lower variance throughout the optimization process, thereby increasing overall stability.
The optimal selection of $\gamma$ should be carefully fine-tuned according to specific application scenarios. We recommend setting the parameter $\gamma$ to around 0.9 based on empirical observations in our experimental context. An appropriate $\gamma$ value strikes a balance between the early-stage exploratory behavior and final convergence.

\subsection{The Benefits of Spectrum Analysis}
\begin{figure*}[t]
    \centering
    \subfigure[Straight line]{
    \label{fig: frequency-linear}
    \includegraphics[width=0.47\linewidth]{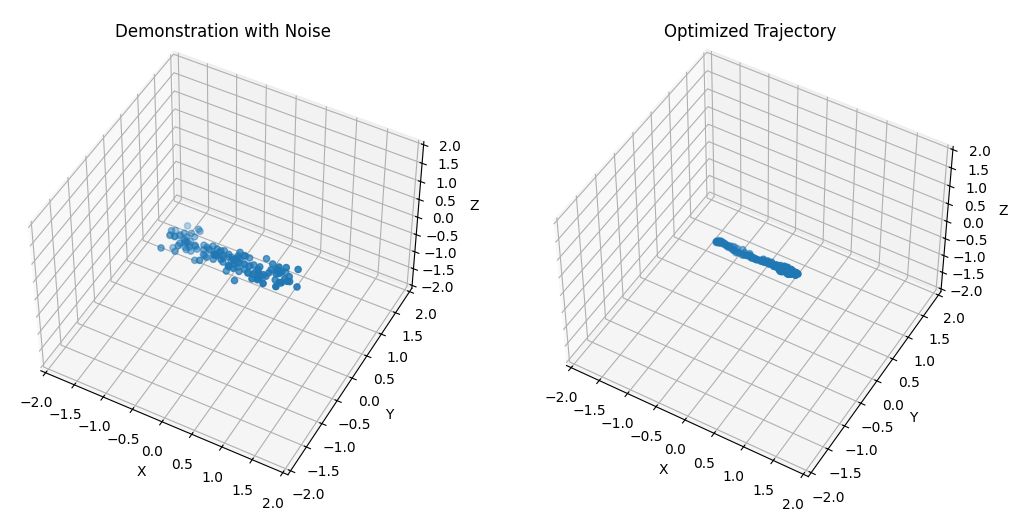}}
    \subfigure[Circle]{
    \label{fig: frequency-round}
    \includegraphics[width=0.47\linewidth]{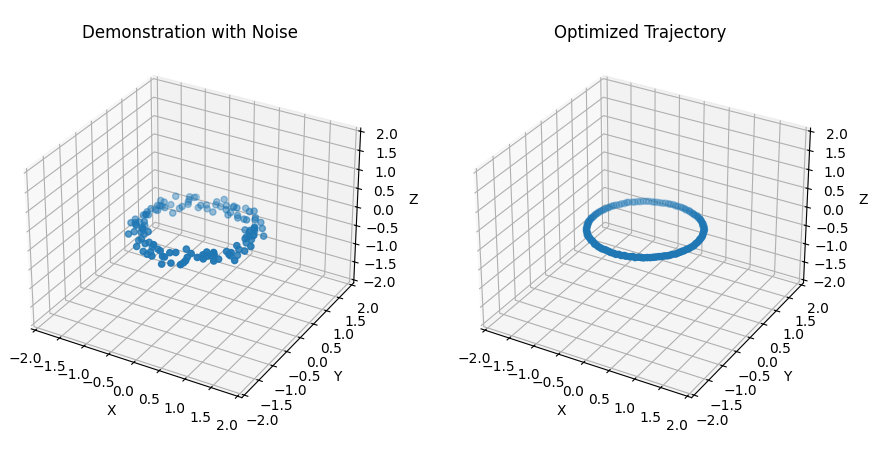}}
    \\
    \subfigure[Semicircle]{
    \label{fig: frequency-back round}
    \includegraphics[width=0.47\linewidth]{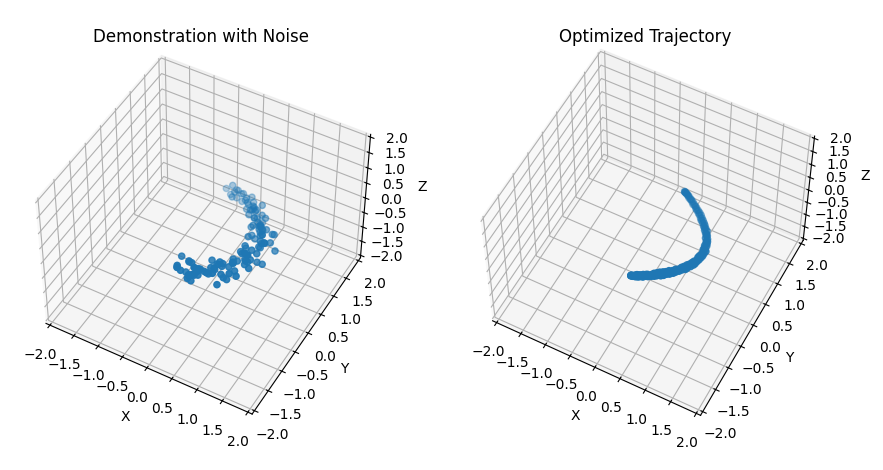}}
    \subfigure[M-shape]{
    \label{fig: frequency-m shape}
    \includegraphics[width=0.47\linewidth]{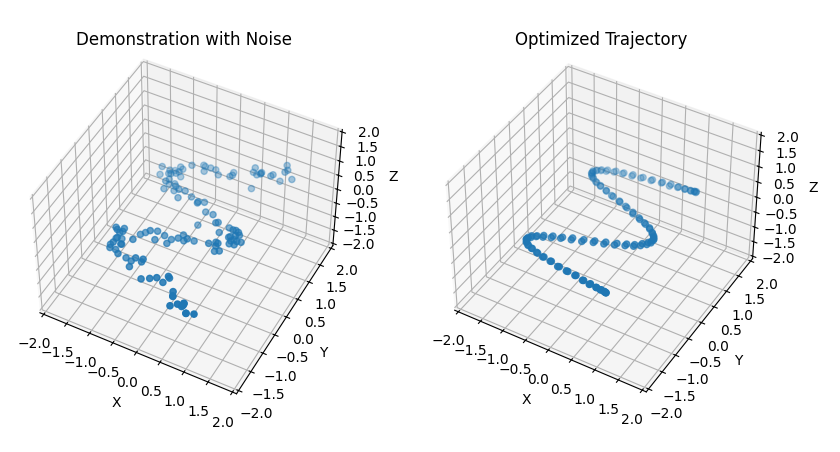}}
    \caption{The denoising performance of the filters in the frequency domain on different trajectories. Different types of trajectories are shown: line in (a), symmetric closed graphic in (b), and unclosed complex graphic in (c) and (d).}
    \label{fig: frequency}
    \vspace{-1em}
\end{figure*}

In this section, we verify the benefits of using spectrum analysis introduced in Section \ref{ss-3.4}. The DFT operation can be seen as an explainable encoder, converting a time-domain sequence into a frequency-domain spectrum to distill out overall trajectory information. Processing a signal in the frequency domain, such as using a filter to remove noise, can improve the overall quality of the signal. By contrast, it is hard to change the overall characteristic of the signal by fine-tuning points in the time domain. We further consider the impact of noise on the various trajectory shapes in the samples and propose a denoising method along with corresponding analyses. The denoising performance of the filters in the frequency domain on different trajectories are shown in \cref{fig: frequency}. 

First, we add the Gaussian noise $\mathcal{N}(\bm{0}, 0.1\bm{I}_3)$ to every point $\bm{p}_i$ in trajectories. After converting them into sequences $\bm{\mathcal{F}}$ in the frequency domain, we employ the gain-controlling method described in \eqref{eq: gain-control} to attenuate frequencies with low amplitudes, converting the noisy line/circular trajectory into a clear one. The constant $\gamma$ can be manually chosen, and we set $\gamma = 20$ in our experiment. The result is shown in~\cref{fig: frequency-linear,fig: frequency-round}.

\begin{equation}\label{eq: gain-control}
    f_{i,k} \leftarrow \frac{f_{i,k}}{\gamma}, \text{if $\lvert f_{i,k}\rvert \leq \gamma$}.
\end{equation}

The above method fails when dealing with non-closed trajectory shapes. For both the semicircle and M-shape trajectories, when processed in the frequency domain, the trajectories tend to be transformed into closed shapes that are distinctly different from their original forms. Past studies have also shown that denoising methods in the frequency domain perform better in closed graphics\cite{li1987shape}. To enhance the gain-controlling method's effectiveness for non-closed curves, we introduce an operation that constructs desired curves by replicating trajectories in the time domain and merging them with the original trajectories through back-stitching. After polishing trajectories in the frequency domain and converting them back to the time domain, we then retain half of the trajectories to eliminate the doubling effect on the trajectory length caused by the previous operation. The polished trajectories of the semicircle and M-shape are shown in \cref{fig: frequency-back round,fig: frequency-m shape}.

We also conduct several experiments where DTW and MSES are used as the imitation cost term, respectively. 
To compare DTW and MSES, we conduct $10$ iterations and calculate the average time consumption and value of imitation costs for each metric, and the results can be seen in Table \ref{tab: MSES}.
The result is consistent with the analysis in Section~\ref{ss-3.4}: the time complexity of MSES is smaller than that of DTW. 
Compared to using DTW as a metric, employing MSES as a metric results in a relatively larger percentage reduction in loss from the initial trajectory values, as shown in the reduction column of Tab.~\ref{tab: MSES}.  According to the time complexity analysis in Section \ref{ss-3.4}, using MSES can quickly minimize the similarity cost term. DTW can be used as a more intuitive time-domain trajectory imitation metric. Based on this, we recommend using MSES to quickly imitate the demonstrated trajectory, followed by DTW for more fine-grained adjustments.

\begin{table}[htbp]
\centering
\caption{Comparison between employing DTW and MSES}
\begin{tabular}{l|ccc|ccc}
\toprule
& \multicolumn{3}{c|}{STOMP} & \multicolumn{3}{c}{MSTOMP} \\
metrics & t/s & value & {reduction} & t/s & value & {reduction}\\
\midrule
DTW & 22.8 & 13.5 & 57.3\%  & 40.6 & 7.8 & {75.3\%} \\
MSES & 7.9 & 0.8 & {92.9\%}  & 14.8 & 0.6 & {94.7\%} \\
\bottomrule
\end{tabular}
\label{tab: MSES}
\vspace{-1em}
\end{table}

\subsection{Performance of MSTOMP in Specific Tasks}
\begin{figure*}[t]
    \centering
    \includegraphics[width=0.9\linewidth]{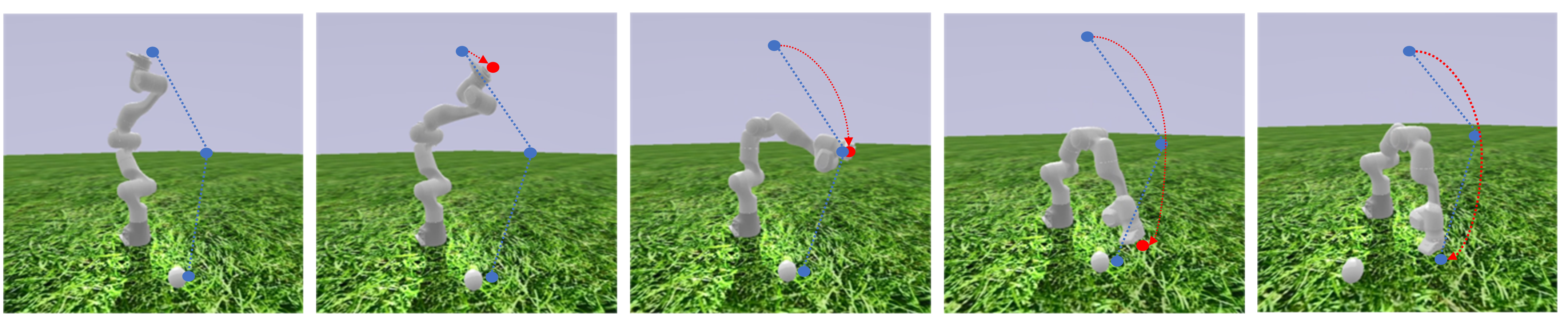}
    \caption{The Panda robotic arm club swings a gold club in the PyBullet environment.}
    \label{fig: experiment1}
    \vspace{-1em}
\end{figure*}
\begin{figure*}[t]
    \centering
    \includegraphics[width=0.7\linewidth]{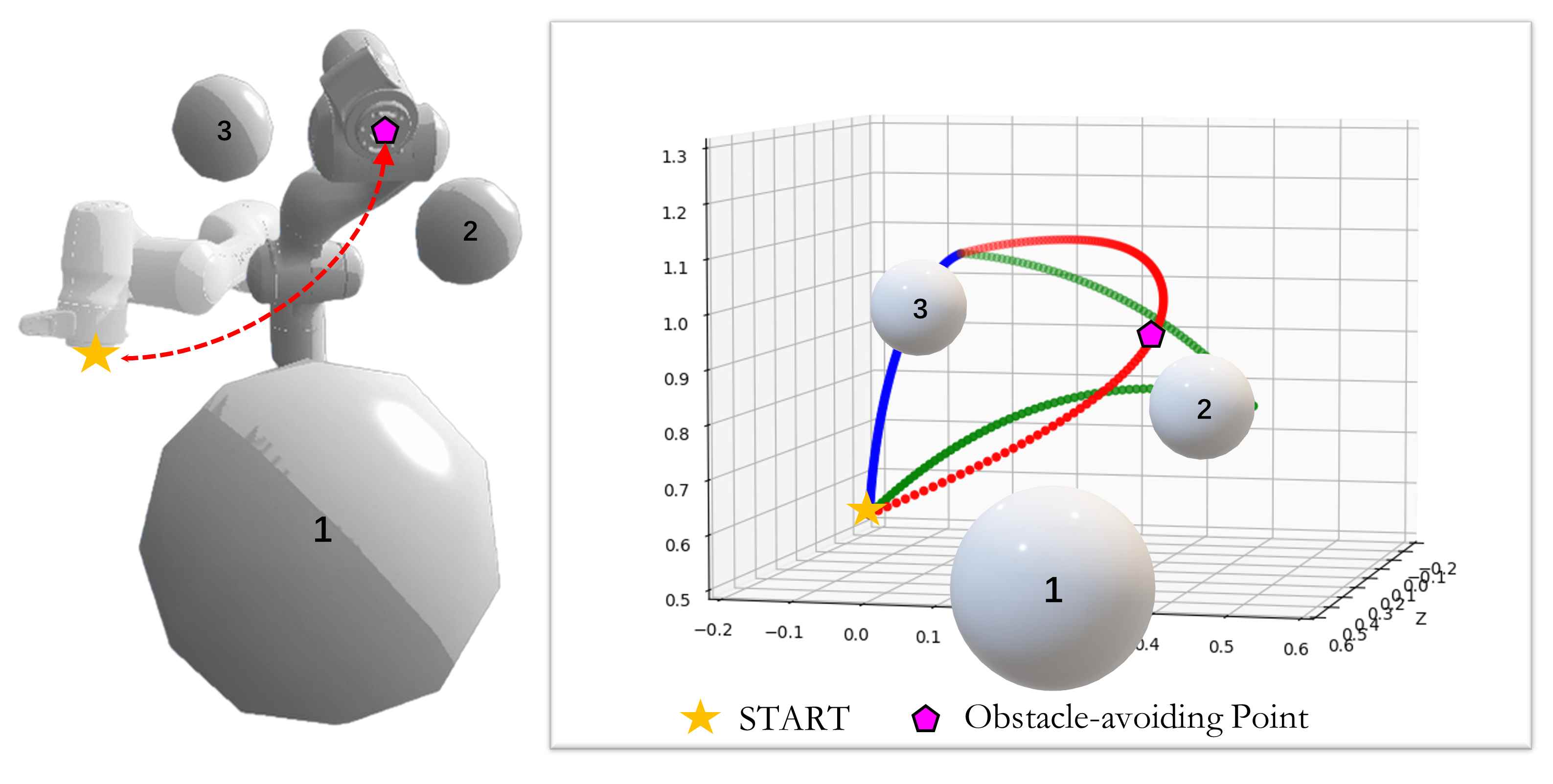}
    \caption{In the obstacle avoidance simulation experiment with the Panda robotic arm, the optimized trajectory moves upwards as a whole to avoid collision, while keeping a safe distance from the right obstacle sphere, even if this results in a higher imitation cost.}
    \label{fig: obs}
    \vspace{-1em}
\end{figure*}
In this part, we implement the MSTOMP in both simulation and the real-world robotic arm. We use the robot to perform specific tasks, e.g., swing the golf club to strike a golf ball. The robot needs to imitate human demonstrations accurately and optimize the club's speed at the precise moment of striking the ball. The simulation results of the club swing process are depicted in \cref{fig: experiment1}. To streamline the model, we directly utilize the end-effector of the robotic arm to strike the ball. Through remote control in the simulator, the robot records a demonstration of a polygonal shape drawn in blue lines. The robot learns the motion trajectory of the swinging club and converts this motion into a joint space sequence. 
From the second to the fifth figures in \cref{fig: experiment1} depict the arc-shaped trajectory of the club swing process of the robotic arm in a red curve, and illustrate the slow-motion capture of a golf club striking the ball. This arc-shaped trajectory enables the robotic arm to have a higher swing speed than that of the polygonal shape trajectory in blue lines. 
{In fact, in the simulation, the swing-hit process unfolds swiftly, resembling the motion executed by an experienced amateur.}

We also test our method's capability for generating trajectories that avoid obstacles. Following the cost function structure in \cite{Stomp}, we calculate the distance between the robotic arm and obstacle spheres and trade it off with $q_d(\bm{\tilde{\theta}}, \bm{\theta}_{demo})$. As shown in \cref{fig: obs}, the robotic arm shifts its trajectory upwards to avoid collision with obstacle spheres and finally reaches the destination point. 
To ensure the safety of the trajectory, when the demonstrated trajectory passes through obstacles in the experimental environment, MSTOMP will guide the robotic arm to generate a new trajectory that bypasses the obstacles with lower collision cost but higher imitation cost $q_d(\bm{\tilde{\theta}}, \bm{\theta}_{demo})$. 
More details on the simulation experiment can be found on \href{https://ming-bot.github.io/MSTOMP.github.io}{our website}. 


\begin{figure}
    \centering
    \includegraphics[width=0.9\linewidth]{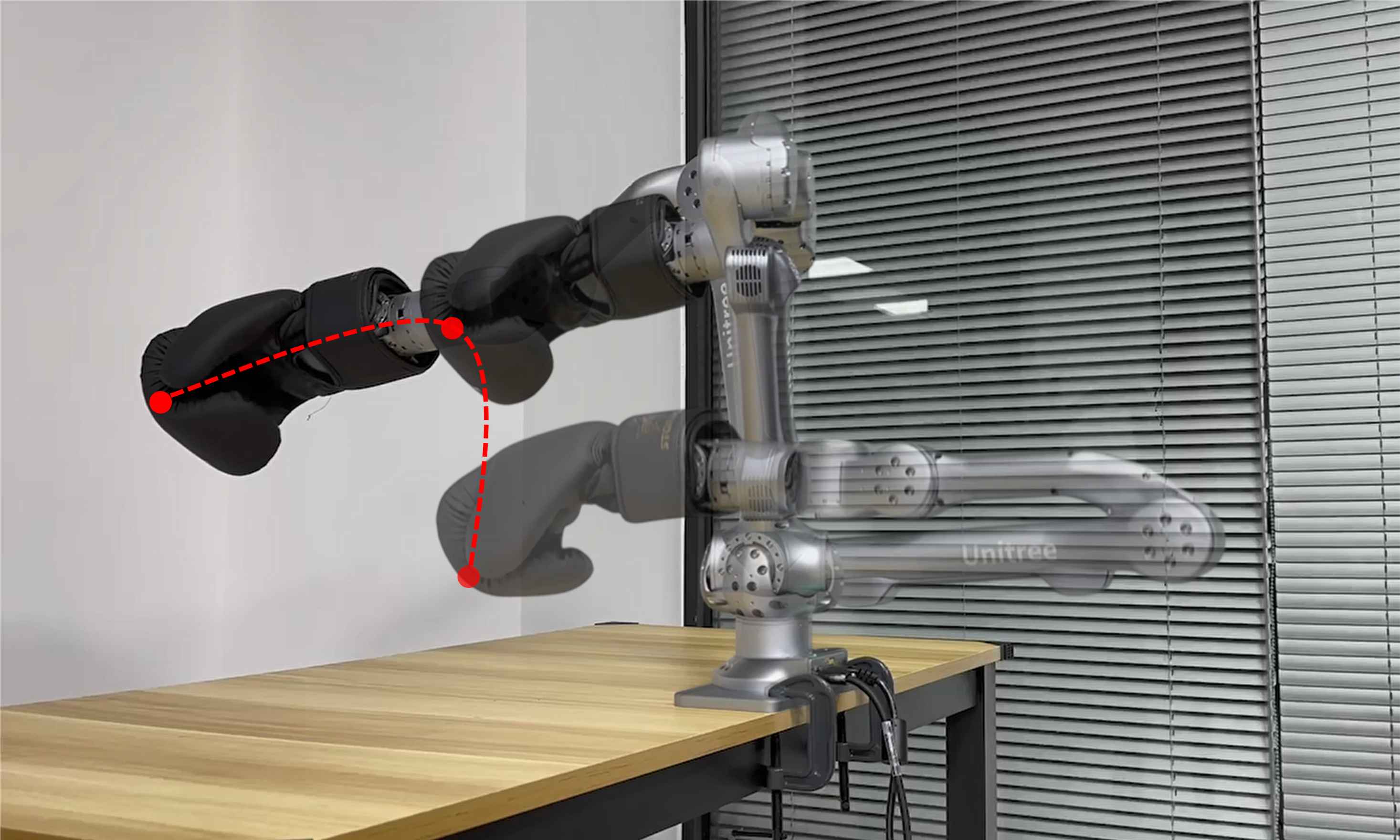}
    \caption{We utilize a Unitree Z1 pro robotic arm to achieve boxing movements. By imitating the slow-motion teaching trajectory and optimizing trajectories with MSTOMP that consider trajectory shape features.}
    \label{fig: boxing}
    \vspace{-1em}
\end{figure}
Then, we conduct real-world experiments on a Unitree Z1 pro robotic arm, as shown in \cref{fig: boxing}. We consider a scenario where the robotic arm attempts to learn how to perform a boxing movement by imitating a human expert.
First, we use the TEACH \& TEACHREPEAT function of the Unitree Z1 pro robotic arm and generate a demonstration trajectory. 
Since the demonstration given by the teaching process can be inefficient (slow, not smooth), we use the proposed MSTOMP algorithm and include terms that represent smoothness and similarity to the cost function to obtain trajectories with better performance. Finally, we execute the optimized trajectory on the robotic arm and present the results in \cref{fig: boxing}.

\section{Conclusion}\label{s-conclusion}
In this paper, we present a novel approach, MSTOMP, for guiding stochastic trajectory optimization to imitate human demonstrations. By leveraging the strengths of STOMP and incorporating two similarity metrics, i.e., DTW and MSES, MSTOMP demonstrates significant improvements in both exploration effectiveness and stability. 
The proposed MSTOMP algorithm is implemented in both simulation and real-world experiments, showcasing its improved optimization performance and stability compared to existing methods. 
Future research directions include learning from multiple demonstrations, enabling the robot to acquire diverse skills and adapt to different scenarios.

\bibliographystyle{IEEEtran}
\balance
\bibliography{b}





\end{document}